\documentclass[11pt,a4paper]{article}
\usepackage[hyperref]{acl2020}
\usepackage{times}
\usepackage{latexsym}

\usepackage{microtype}

\aclfinalcopy 
\def\aclpaperid{1651} 

\usepackage{url}
\usepackage{hyperref}
\usepackage{booktabs}
\usepackage{graphicx}
\usepackage{subfigure}
\usepackage{bm}
\usepackage{color}
\usepackage{amsmath}
\usepackage{amsfonts}
\usepackage{amssymb}
\usepackage{stmaryrd}
\usepackage{stfloats}
\usepackage{xspace}

\newcommand{\ie}{\emph{i.e.,}\xspace}
\newcommand{\eg}{\emph{e.g.,}\xspace}

\title{Span-based Localizing Network for Natural Language Video Localization}

\author{
   Hao Zhang$^{1,2}$,~~Aixin Sun$^{1}$,~~Wei Jing$^{2,3}$,~~Joey Tianyi Zhou$^{2,}$\thanks{$\;\;$Corresponding author.} \\
   $^{1}$School of Computer Science and Engineering, Nanyang Technological University, Singapore\\
   $^{2}$Institute of High Performance Computing, A*STAR, Singapore\\
   $^{3}$Institute for Infocomm Research, A*STAR, Singapore\\
   \texttt{hao007@e.ntu.edu.sg,}~~
   \texttt{axsun@ntu.edu.sg}\\
   \texttt{21wjing@gmail.com,}~~
   \texttt{joey\_zhou@ihpc.a-star.edu.sg}
}

\date{}

\begin{document}
\maketitle
\begin{abstract}
Given an untrimmed video and a text query, natural language video localization (NLVL) is to locate a matching span from the video that semantically corresponds to the query. Existing solutions formulate NLVL either as a ranking task and apply multimodal matching architecture, or as a regression task to directly regress the target video span. In this work, we address NLVL task with a span-based QA approach by treating the input video as text passage. We propose a video span localizing network (VSLNet), on top of the standard span-based QA framework, to address NLVL. The proposed VSLNet tackles the differences between NLVL and span-based QA through a simple yet effective query-guided highlighting (QGH) strategy. The QGH guides VSLNet to search for matching video span within a highlighted region. Through extensive experiments on three benchmark datasets, we show that the proposed VSLNet outperforms the state-of-the-art methods; and adopting span-based QA framework is a promising direction to solve NLVL.\footnote{https://github.com/IsaacChanghau/VSLNet} 
\end{abstract}

\section{Introduction}\label{sec:intro}

\begin{figure}[t]
    \centering
    \includegraphics[width=0.48\textwidth]{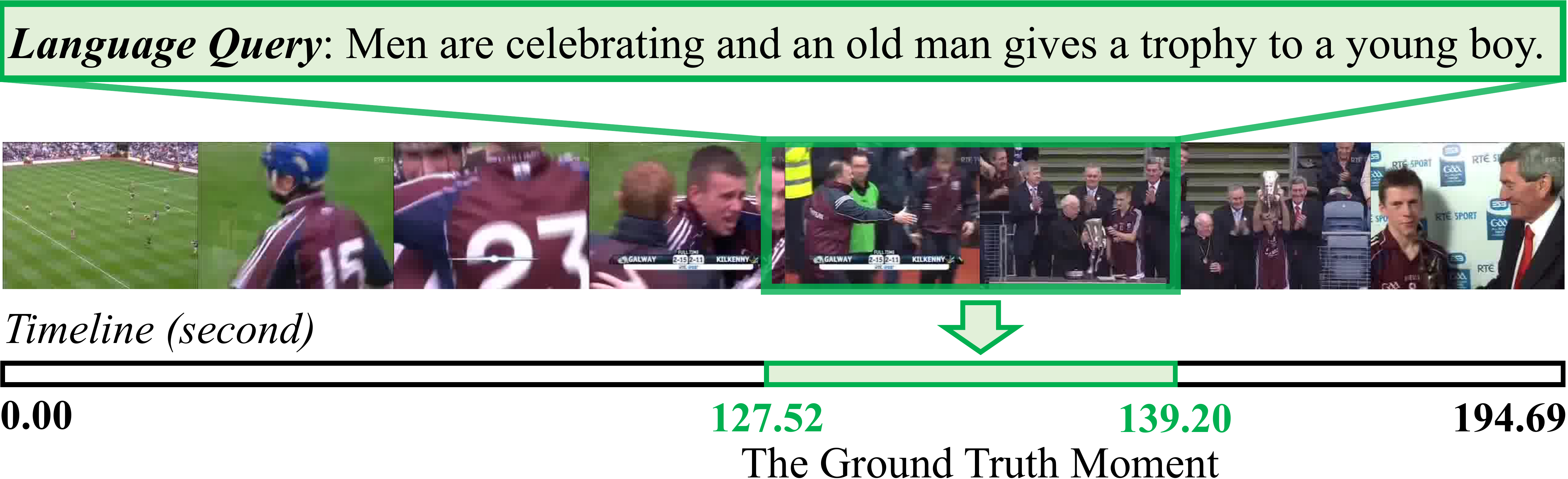}
    \caption{\small An illustration of localizing a temporal moment in an untrimmed video by a given language query.}
	\label{fig_example}
\end{figure}

Given an untrimmed video, natural language video localization (NLVL) is to retrieve or localize a temporal moment that semantically corresponds to a given language query. An example is shown in  Figure~\ref{fig_example}. As an important vision-language understanding task, NLVL involves both computer vision and natural language processing techniques~\cite{krishna2017dense,Hendricks2017LocalizingMI,Gao2018MotionAppearanceCN,le2019multimodal,yu2019activityqa}. Clearly, cross-modal reasoning is essential for NLVL to correctly locate the target moment from a video.

Prior works primarily treat NLVL as a ranking task, which is solved by applying multimodal matching architecture to find the best matching video segment for a given language query~\cite{Gao2017TALLTA,hendricks2018localizing,Liu2018AMR,ge2019mac,Xu2019MultilevelLA,chen2019semantic,zhang2019man}. Recently, some works explore to model cross-interactions between video and query, and to regress the temporal locations of target moment directly~\cite{Yuan2019ToFW,lu2019debug}. There are also studies to formulate NLVL as a sequence decision making problem and to solve it by reinforcement learning~\cite{Wang2019LanguageDrivenTA,he2019Readwa}.

We address the NLVL task from a different perspective. The essence of NLVL is to search for a video moment as the answer to a given language query from an untrimmed video.  By treating the video as a text passage, and the target moment as the answer span, NLVL shares significant similarities with span-based question answering (QA) task. The span-based QA framework~\cite{Seo2017BidirectionalAF,wang2017gated,huang2018fusionnet} can be adopted for NLVL. Hence, we attempt to solve this task with a multimodal span-based QA approach.

There are two main differences between traditional text span-based QA and NLVL tasks. First, video is continuous and causal relations between video events are usually adjacent. Natural language, on the other hand, is inconsecutive and words in a sentence demonstrate syntactic structure. For instance, changes between adjacent video frames are usually very small, while adjacent word tokens may carry distinctive meanings. As the result, many events in a video are directly correlated and can even cause one another~\cite{krishna2017dense}. Causalities between word spans or sentences are usually indirect and can be far apart. Second, compared to word spans in text, human is insensitive to small shifting between video frames. In other words,  small offsets between video frames do not affect the understanding of video content, but the differences of a few words or even one word could change the meaning of a sentence.

As a baseline, we first solve the NLVL task with a standard span-based QA framework named \textbf{VSLBase}. Specifically, visual features are analogous to that of text passage; the target moment is regarded as the answer span. VSLBase is trained to predict the start and end boundaries of the answer span. Note that VSLBase does not address the two aforementioned major differences between video and natural language. To this end, we propose an improved version named \textbf{VSLNet} (Video Span Localizing Network). VSLNet introduces a Query-Guided Highlighting (\textbf{QGH}) strategy in addition to VSLBase. Here, we regard the target moment and its adjacent contexts as foreground, while the rest as background, \ie foreground covers a slightly longer span than the answer span.  With QGH, VSLNet is guided to search for the target moment within a highlighted region. Through region highlighting, VSLNet well addresses the two differences. First, the longer region provides additional contexts for locating answer span due to the  continuous nature of video content. Second, the highlighted region helps the network to focus on subtle differences between video frames, because the search space is reduced compared to the full video. 

Experimental results on three benchmark datasets show that adopting span-based QA framework is suitable for NLVL. With a simple network architecture, VSLBase delivers comparable performance to strong baselines. In addition, VSLNet further boosts the performance and achieves the best among all evaluated methods.

\section{Related Work}\label{sec:related}

\paragraph{Natural Language Video Localization.} The task of retrieving video segments using language queries was introduced in~\cite{Hendricks2017LocalizingMI,Gao2017TALLTA}. Solutions to NLVL need to model the cross-interactions between natural language and video. The early works treat NLVL as a ranking task, and rely on multimodal matching architecture to find the best matching video moment for a language query~\cite{Gao2017TALLTA,Hendricks2017LocalizingMI,hendricks2018localizing,wu2018multimodal,Liu2018AMR,Liu2018CML,Xu2019MultilevelLA,zhang2019man}. Although intuitive, these models are sensitive to negative samples. Specifically, they need to dense sample candidate moments to achieve good performance, which leads to low efficiency and lack of flexibility.

Various approaches have been proposed to overcome those drawbacks. \citet{Yuan2019ToFW} builds a proposal-free method using BiLSTM and directly regresses temporal locations of target moment. \citet{lu2019debug} proposes a dense bottom-up framework, which regresses the distances to start and end boundaries for each frame in target moment, and select the ones with highest confidence as final result. \citet{yuan2019semantic} proposes a semantic conditioned dynamic modulation for better correlating sentence related video contents over time, and establishing a precise matching relationship between sentence and video. There are also works~\cite{Wang2019LanguageDrivenTA,he2019Readwa} that formulate NLVL as a sequence decision making problem, and adopt reinforcement learning based approaches, to progressively observe candidate moments conditioned on language query.

Most similar to our work are~\cite{chen2019localizing} and~\cite{ghosh2019excl}, as both studies are considered using the concept of question answering to address NLVL. However, both studies do not explain the similarity and differences between NLVL and traditional span-based QA, and they do not adopt the standard span-based QA framework. In our study, VSLBase adopts standard span-based QA framework; and VSLNet explicitly addresses the differences between NLVL and traditional span-based QA tasks.

\paragraph{Span-based Question Answering.} 
Span-based QA has been widely studied in past years. \citet{wang2017machineCU} combines match-LSTM~\cite{wang2016learning} and Pointer-Net~\cite{vinyals2015pointer} to estimate boundaries of the answer span. BiDAF~\cite{Seo2017BidirectionalAF} introduces bi-directional attention to obtain query-aware context representation. \citet{Xiong2017DynamicCN} proposes a coattention network to capture the interactions between context and query. R-Net~\cite{wang2017gated} integrates mutual and self attentions into RNN encoder for feature refinement. QANet~\cite{wei2018fast} leverages a similar attention mechanism in a stacked convolutional encoder to improve performance. FusionNet~\cite{huang2018fusionnet} presents a full-aware multi-level attention to capture complete query information. By treating input video as text passage, the above frameworks are all applicable to NLVL in principle. However, these frameworks are not designed to consider the differences between video and text passage. Their modeling complexity arises from the interactions between query and text passage, both are text. In our solution, VSLBase adopts a simple and standard span-based QA framework, making it easier to model the differences between video and text through adding additional modules. Our VSLNet addresses the differences by introducing the QGH module.

Very recently, pre-trained transformer based language models~\cite{devlin2019bert,dai2019transformer,liu2019roberta,yang2019xlNet} have elevated the performance of span-based QA tasks by a large margin. Meanwhile, similar pre-trained models~\cite{sun2019contrastive,sun2019videobert,yu2019adapting,rahman2019mbert,nguyen2019multi,lu2019vilbert,tan2019lxmert} are being proposed to learn joint distributions over multimodality sequence of visual and linguistic inputs. Exploring the pre-trained models for NLVL is part of our future work and is out of the scope of this study.

\begin{figure*}[t]
    \centering
    \includegraphics[width=1\textwidth]{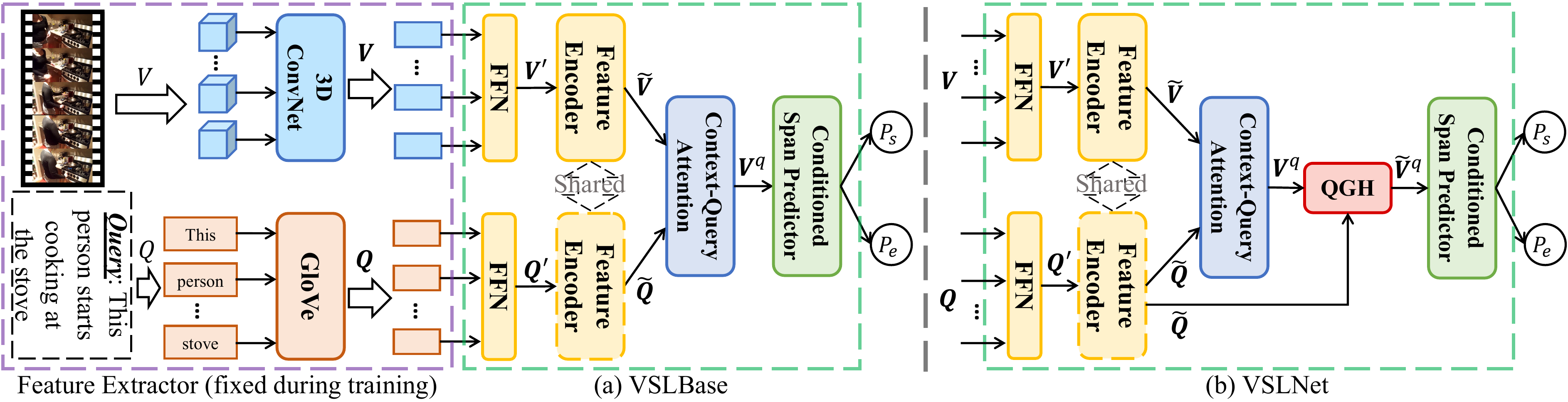}
    \caption{\small An overview of the proposed architecture for NLVL. The feature extractor is fixed during training. Figure (a) depicts the adoption of standard span-based QA framework, \ie VSLBase. Figure (b) shows the structure of VSLNet.}
	\label{fig_architecture}
\end{figure*}

\section{Methodology}\label{sec:method}
We now describe how to address NLVL task by adopting a span-based QA framework. We then present VSLBase (Sections~\ref{ssec:feature} to~\ref{ssec:spanpred}) and VSLNet in detail. Their architectures are shown in Figure~\ref{fig_architecture}.

\subsection{Span-based QA for NLVL}
We denote the untrimmed video as $V=\{f_t\}_{t=1}^{T}$ and the language query as $Q=\{q_j\}_{j=1}^{m}$, where $T$ and $m$ are the number of frames and words, respectively. $\tau^{s}$ and $\tau^{e}$ represent the start and end time of the temporal moment \ie answer span. To address NLVL with span-based QA framework, its data is transformed into a set of SQuAD style triples $(Context, Question, Answer)$~\cite{rajpurkar2016squad}. For each video $V$, we extract its visual features $\mathbf{V}=\{\mathbf{v}_i\}_{i=1}^{n}$ by a pre-trained 3D ConvNet~\cite{Carreira2017QuoVA}, where $n$ is the number of extracted features. Here, $\mathbf{V}$ can be regarded as the sequence of word embeddings for a text passage with $n$ tokens. Similar to word embeddings, each feature $\mathbf{v}_i$ here is a video feature vector.

Since span-based QA aims to predict start and end boundaries of an answer span, the start/end time of a video sequence needs to be mapped to the corresponding boundaries in the visual feature sequence $\mathbf{V}$. Suppose the video duration is $\mathcal{T}$, the start (end) span index is calculated by $a^{s(e)}=\langle\tau^{s(e)}/\mathcal{T}\times n\rangle$, where $\langle\cdot\rangle$ denotes the rounding operator. During the inference, the predicted span boundary can be easily converted to the corresponding time via $\tau^{s(e)}=a^{s(e)}/n\times \mathcal{T}$.

After transforming moment annotations in NLVL dataset, we obtain a set of $(\mathbf{V},Q,\mathbf{A})$ triples. Visual features $\mathbf{V}=[\mathbf{v}_1, \mathbf{v}_2, \dots, \mathbf{v}_n]$ act as the passage with $n$ tokens; $Q=[q_1, q_2,\dots, q_m]$ is the query with $m$ tokens, and the answer $\mathbf{A}=[\mathbf{v}_{a^{s}}, \mathbf{v}_{a^{s}+1}, \dots, \mathbf{v}_{a^{e}}]$ corresponds to a piece in the passage. Then, the NLVL task becomes to find the correct start and end boundaries of the answer span, $a^{s}$ and $a^{e}$.

\subsection{Feature Encoder}\label{ssec:feature}
We already have visual features $\mathbf{V}=\{\mathbf{v}_i\}_{i=1}^{n}\in\mathbb{R}^{n\times d_v}$. Word embeddings of a text query $Q$, $\mathbf{Q}=\{\mathbf{q}_j\}_{j=1}^{m}\in\mathbb{R}^{m\times d_q}$, are easily obtainable \eg GloVe. We project them into the same dimension $d$, $\mathbf{V'}\in\mathbb{R}^{n\times d}$ and $\mathbf{Q'}\in\mathbb{R}^{m\times d}$, by two linear layers (see Figure~\ref{fig_architecture}(a)). Then we build the feature encoder with a simplified version of the embedding encoder layer in QANet~\cite{wei2018fast}.

Instead of applying a stack of multiple encoder blocks, we use only one encoder block. This encoder block consists of four convolution layers, followed by a multi-head attention layer~\cite{vaswani2017attention}. A feed-forward layer is used to produce the output. Layer normalization~\cite{Ba2016LayerN} and residual connection~\cite{he2016resnet} are applied to each layer. The encoded visual features and word embeddings are as follows:
\begin{equation}
\begin{aligned}
    \mathbf{\widetilde{V}} & =\mathtt{FeatureEncoder}(\mathbf{V'}) \\
    \mathbf{\widetilde{Q}} & =\mathtt{FeatureEncoder}(\mathbf{Q'})
\end{aligned}
\end{equation}
The parameters of feature encoder are shared by visual features and word embeddings.

\subsection{Context-Query Attention}\label{ssec:attention}
After feature encoding, we use context-query attention (CQA)~\cite{Seo2017BidirectionalAF,Xiong2017DynamicCN,wei2018fast} to capture the cross-modal interactions between visual and textural features. CQA first calculates the similarity scores, $\mathcal{S}\in\mathbb{R}^{n\times m}$, between each visual feature and query feature. Then context-to-query ($\mathcal{A}$) and query-to-context ($\mathcal{B}$) attention weights are computed as:\begin{equation}
    \mathcal{A}=\mathcal{S}_{r}\cdot\mathbf{\widetilde{Q}}\in\mathbb{R}^{n\times d}, \mathcal{B}=\mathcal{S}_{r}\cdot\mathcal{S}_{c}^{T}\cdot\mathbf{\widetilde{V}}\in\mathbb{R}^{n\times d}\nonumber
\end{equation}
where $\mathcal{S}_{r}$ and $\mathcal{S}_{c}$ are the row- and column-wise normalization of $\mathcal{S}$ by SoftMax, respectively. Finally, the output of context-query attention is written as:
\begin{equation}
    \mathbf{V}^{q}=\mathtt{FFN}\big([\mathbf{\widetilde{V}};\mathcal{A};\mathbf{\widetilde{V}}\odot\mathcal{A};\mathbf{\widetilde{V}}\odot\mathcal{B}]\big)
\end{equation}
where $\mathbf{V}^{q}\in\mathbb{R}^{n\times d}$; $\mathtt{FFN}$ is a single feed-forward layer; $\odot$ denotes element-wise multiplication.

\subsection{Conditioned Span Predictor}\label{ssec:spanpred}
We construct a conditioned span predictor by using two unidirectional LSTMs and two feed-forward layers, inspired by~\citet{ghosh2019excl}. The main difference between ours and~\citet{ghosh2019excl} is that we use unidirectional LSTM instead of bidirectional LSTM. We observe that unidirectional LSTM shows similar performance with fewer parameters and higher efficiency. The two LSTMs are stacked so that the LSTM of end boundary can be conditioned on that of start boundary.
Then the hidden states of the two LSTMs are fed into the corresponding feed-forward layers to compute the start and end scores:
\begin{equation}
\begin{aligned}
    \mathbf{h}_{t}^{s} & =\mathtt{UniLSTM}_\textrm{start}(\mathbf{v}_{t}^{q}, \mathbf{h}_{t-1}^{s}) \\
    \mathbf{h}_{t}^{e} & =\mathtt{UniLSTM}_\textrm{end}(\mathbf{h}_{t}^{s}, \mathbf{h}_{t-1}^{e}) \\
    \mathbf{S}_{t}^{s} & = \mathbf{W}_{s}\times([\mathbf{h}_{t}^{s};\mathbf{v}_{t}^{q}]) + \mathbf{b}_{s} \\
    \mathbf{S}_{t}^{e} & = \mathbf{W}_{e}\times([\mathbf{h}_{t}^{e};\mathbf{v}_{t}^{q}]) + \mathbf{b}_{e}
\end{aligned}\label{eq_predictor}
\end{equation}
Here, $\mathbf{S}_{t}^{s}$ and $\mathbf{S}_{t}^{e}$ denote the scores of start and end boundaries at position $t$; $\mathbf{v}_{t}^{q}$ represents the $t$-th feature in $\mathbf{V}^{q}$. $\mathbf{W}_{s/e}$ and $\mathbf{b}_{s/e}$ denote the weight matrix and bias of the start/end feed-forward layer, respectively.
Then, the probability distributions of start and end boundaries are computed by $P_{s} = \textrm{SoftMax}(\mathbf{S}^{s})\in\mathbb{R}^{n}$ and $P_{e} = \textrm{SoftMax}(\mathbf{S}^{e})\in\mathbb{R}^{n}$, and the training objective is defined as:
\begin{equation}
    \mathcal{L}_{\textrm{span}} = \frac{1}{2}\big[f_{\textrm{CE}}(P_{s}, Y_{s}) + f_{\textrm{CE}}(P_{e}, Y_{e})\big]
\end{equation}
where $f_{\textrm{CE}}$ represents cross-entropy loss function; $Y_{s}$ and $Y_{e}$ are the labels for the start ($a^{s}$) and end ($a^{e}$) boundaries, respectively. During inference, the predicted answer span $(\hat{a}^{s},\hat{a}^{e})$ of a query is generated by maximizing the joint probability of start and end boundaries by:
\begin{equation}
\begin{aligned}
    \mathtt{span}(\hat{a}^{s},\hat{a}^{e}) & = \arg\max_{\hat{a}^{s},\hat{a}^{e}} P_{s}(\hat{a}^{s}) P_{e}(\hat{a}^{e}) \\
    & \textrm{s.t. } 0\leq\hat{a}^{s}\leq \hat{a}^{e}\leq n
\end{aligned}
\end{equation}

We have completed the VSLBase architecture (see Figure~\ref{fig_architecture}(a)). VSLNet is built on top of VSLBase with QGH, to be detailed next. 

\begin{figure}[t]
    \centering
    \includegraphics[width=0.48\textwidth]{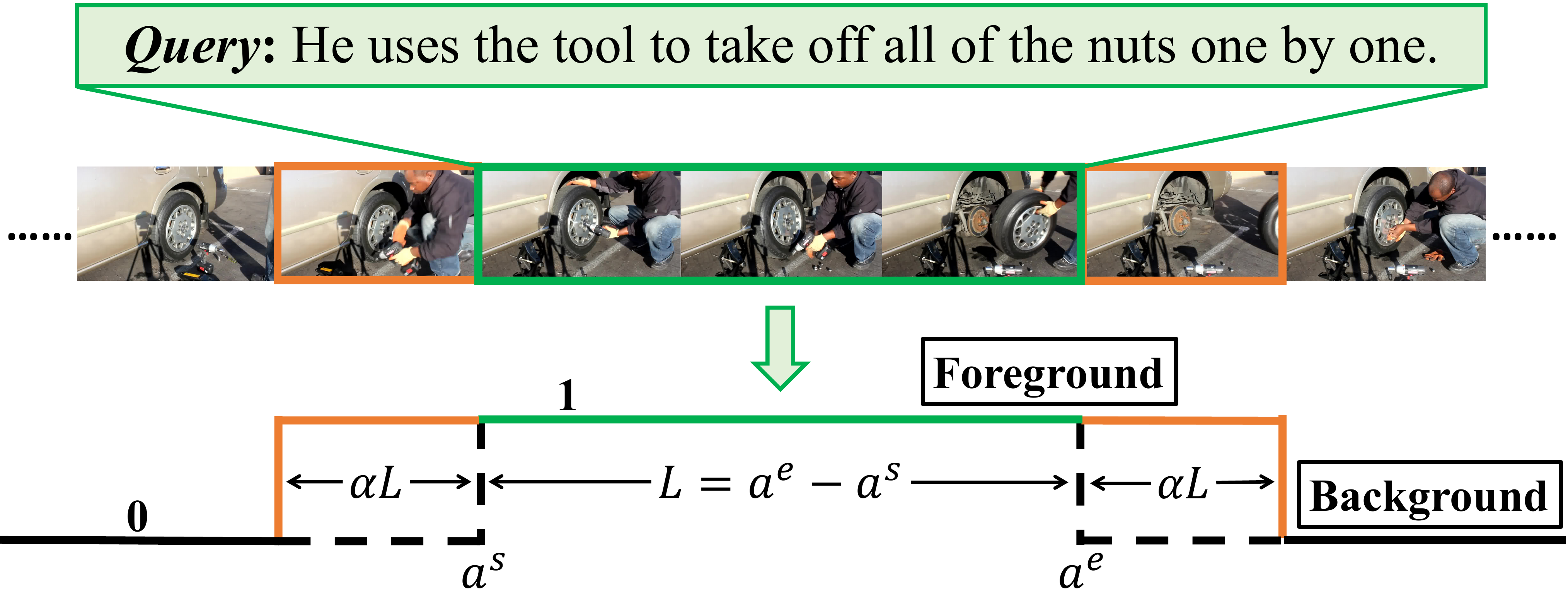}
    \caption{\small An illustration of foreground and background of visual features. $\alpha$ is the ratio of foreground extension.}
	\label{fig_mask}
\end{figure}

\subsection{Query-Guided Highlighting}\label{ssec:qgh}
A Query-Guided Highlighting (QGH) strategy is introduced in VSLNet, to address the major differences between text span-based QA and NLVL tasks, as shown in Figure~\ref{fig_architecture}(b). With QGH strategy, we consider the target moment as the foreground, and the rest as background, illustrated in Figure~\ref{fig_mask}. The target moment, which is aligned with the language query, starts from $a^{s}$ and ends at $a^{e}$ with length $L=a^{e}-a^{s}$. QGH extends the boundaries of the foreground to cover its antecedent and consequent video contents, where the extension ratio is controlled by a hyperparameter $\alpha$. As aforementioned in Introduction, the extended boundary could potentially cover additional contexts and also help the network to focus on subtle differences between video frames. 

\begin{figure}[t]
    \centering
    \includegraphics[width=0.45\textwidth]{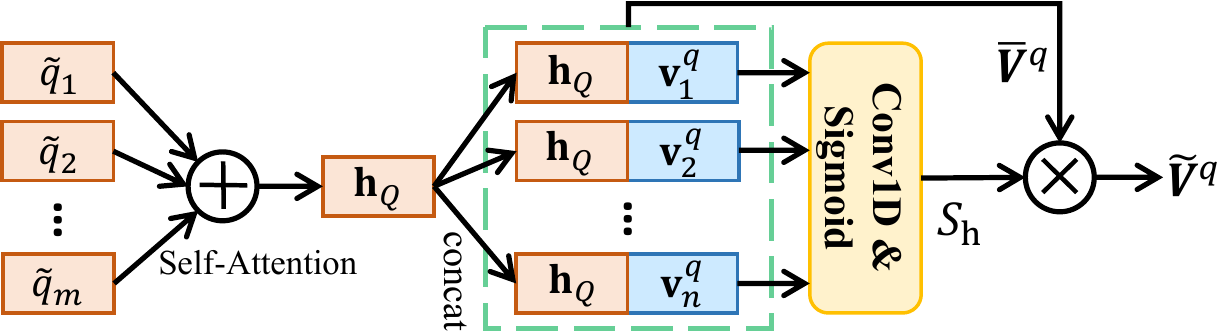}
    \caption{\small The structure of Query-Guided Highlighting.}
	\label{fig_qgm}
\end{figure}

By assigning $1$ to foreground and $0$ to background, we obtain a sequence of $0$-$1$, denoted by $Y_{\textrm{h}}$. QGH is a binary classification module to predict the confidence a visual feature belongs to foreground or background.
The structure of QGH is shown in Figure~\ref{fig_qgm}. We first encode  word features $\mathbf{\widetilde{Q}}$ into sentence representation (denoted by $\mathbf{h}_{Q}$), with self-attention mechanism~\cite{Bahdanau2015NeuralMT}.
Then $\mathbf{h}_{Q}$ is concatenated with each feature in $\mathbf{V}^{q}$ as $\mathbf{\bar{V}}^{q}=[\mathbf{\bar{v}}_{1}^{q},\dots,\mathbf{\bar{v}}_{n}^{q}]$, where $\mathbf{\bar{v}}_{i}^{q}=[\mathbf{v}_{i}^{q};\mathbf{h}_{Q}]$.
The highlighting score is computed as:
\begin{equation}
    \mathcal{S}_\textrm{h}=\sigma\big(\mathtt{Conv1D}(\mathbf{\bar{V}}^{q})\big)\nonumber
\end{equation}
where $\sigma$ denotes Sigmoid activation; $\mathcal{S}_{\textrm{h}}\in\mathbb{R}^{n}$. The highlighted features are calculated by:
\begin{equation}
    \mathbf{\widetilde{V}}^{q}=\mathcal{S}_\textrm{h}\cdot\mathbf{\bar{V}}^{q}
\end{equation}

Accordingly, feature $\mathbf{V}^{q}$ in Equation~\ref{eq_predictor} is replaced by $\mathbf{\widetilde{V}}^{q}$ in VSLNet to compute  $\mathcal{L}_{\textrm{span}}$. The loss function of query-guided highlighting is formulated as:
\begin{equation}
    \mathcal{L}_{\textrm{QGH}}=f_{\textrm{CE}}(S_{\textrm{h}}, Y_{\textrm{h}})
\end{equation}
VSLNet is trained in an end-to-end manner by minimizing the following loss:
\begin{equation}
    \mathcal{L}=\mathcal{L}_{\textrm{span}} + \mathcal{L}_{\textrm{QGH}}.
\end{equation}

\section{Experiments}\label{sec:exp}

\subsection{Datasets}
We conduct experiments on three benchmark datasets: Charades-STA~\cite{Gao2017TALLTA}, ActivityNet Caption~\cite{krishna2017dense}, and TACoS~\cite{regneri2013grounding}, summarized in Table~\ref{tab-data}.

\textbf{Charades-STA} is prepared by \citet{Gao2017TALLTA} based on Charades dataset \cite{sigurdsson2016hollywood}. The videos are  about daily indoor activities. There are $12,408$ and $3,720$ moment annotations for training and test, respectively.

\textbf{ActivityNet Caption} contains about $20$k videos
taken from ActivityNet~\cite{heilbron2015activitynet}. We follow the setup in~\citet{Yuan2019ToFW}, leading to $37,421$ moment annotations for training, and $17,505$ annotations for test.

\textbf{TACoS} is selected from MPII Cooking Composite Activities dataset~\cite{Rohrbach2012SDA}. We follow the setting in~\citet{Gao2017TALLTA}, where $10,146$, $4,589$ and $4,083$ annotations are used for training, validation and test, respectively.

\begin{table*}[t]
    \small
    \setlength{\tabcolsep}{2.2pt}
	\centering
	\begin{tabular}{ l c c c r r r r r }
		\toprule
		Dataset & Domain & \# Videos (train/val/test) & \# Annotations & $N_{\textrm{vocab}}$ & $\bar{L}_{video}$ & $\bar{L}_{query}$ & $\bar{L}_{moment}$ & $\Delta_{moment}$  \\
		\midrule
        Charades-STA & Indoors & $5,338/-/1,334$ & $12,408/-/3,720$ & $1,303$ & $30.59s$ & $7.22$ & $8.22s$ & $3.59s$ \\
        \midrule
        ActivityNet Cap & Open & $10,009/-/4,917$ & $37,421/-/17,505$ & $12,460$ & $117.61s$ & $14.78$ & $36.18s$ & $40.18s$ \\
        \midrule
        TACoS & Cooking & $75/27/25$ & $10,146/4,589/4,083$ & $2,033$ & $287.14s$ & $10.05$ & $5.45s$ & $7.56s$ \\
        \bottomrule
	\end{tabular}
	\caption{\small Statistics of NLVL datasets, where $N_{\textrm{vocab}}$ is vocabulary size of lowercase words, $\bar{L}_{video}$ denotes average length of videos in seconds, $\bar{L}_{query}$ denotes average number of words in sentence query, $\bar{L}_{moment}$ is average length of temporal moments in seconds, and $\Delta_{moment}$ is the standard deviation of temporal moment length in seconds.}
	\label{tab-data}
\end{table*}

\subsection{Experimental Settings}
\paragraph{Metrics.}
We adopt ``$\textrm{R@}n, \textrm{IoU}=\mu$'' and ``mIoU'' as the evaluation metrics, following~\cite{Gao2017TALLTA,Liu2018AMR,Yuan2019ToFW}. The ``$\textrm{R@}n, \textrm{IoU}=\mu$'' denotes the percentage of language queries having at least one result whose Intersection over Union (IoU) with ground truth is larger than $\mu$ in top-\textit{n} retrieved moments. ``mIoU'' is the average IoU over all testing samples. In our experiments, we use $n=1$ and $\mu\in\{0.3, 0.5, 0.7\}$.

\paragraph{Implementation.}
For language query $Q$, we use $300$d GloVe~\cite{pennington2014glove} vectors to initialize each lowercase word; the word embeddings are fixed during training.
For untrimmed video $V$, we downsample frames and extract RGB visual features using the 3D ConvNet which was pre-trained on Kinetics dataset~\cite{Carreira2017QuoVA}.
We set the dimension of all the hidden layers in the model as 128; the kernel size of convolution layer is $7$; the head size of multi-head attention is $8$. For all datasets, the model is trained for $100$ epochs with batch size of $16$ and early stopping strategy. Parameter optimization is performed by Adam~\cite{Kingma2015AdamAM} with learning rate of $0.0001$, linear decay of learning rate and gradient clipping of $1.0$. Dropout~\cite{srivastava2014dropout} of $0.2$ is applied to prevent overfitting.

\subsection{Comparison with State-of-the-Arts}
We compare VSLBase and VSLNet with the following state-of-the-arts: CTRL~\cite{Gao2017TALLTA}, ACRN~\cite{Liu2018AMR}, TGN~\cite{chen2018temporally}, ACL-K~\cite{ge2019mac}, QSPN~\cite{Xu2019MultilevelLA}, SAP~\cite{chen2019semantic}, MAN~\cite{zhang2019man}, SM-RL~\cite{Wang2019LanguageDrivenTA}, RWM-RL~\cite{he2019Readwa}, L-Net~\cite{chen2019localizing}, ExCL~\cite{ghosh2019excl}, ABLR~\cite{Yuan2019ToFW} and DEBUG~\cite{lu2019debug}. In all result tables, the scores of compared methods are reported in the corresponding works. Best results are in \textbf{bold} and second best \underline{underlined}. 

\begin{table}[t]
   \small
	\centering
    \setlength{\tabcolsep}{3.0 pt}
	\begin{tabular}{l c c c c}
		\toprule
        Model & $\textrm{IoU}=0.3$ & $\textrm{IoU}=0.5$ & $\textrm{IoU}=0.7$ & mIoU \\
        \midrule
        \multicolumn{5}{c}{3D ConvNet without fine-tuning as visual feature extractor } \\
        \hline
        CTRL & - & $23.63$ & $8.89$  & - \\
        ACL-K & - & $30.48$ & $12.20$ & - \\
        QSPN & $54.70$ & $35.60$ & $15.80$ & - \\
        SAP & - & $27.42$ & $13.36$ & - \\
        SM-RL & - & $24.36$ & $11.17$ & - \\
        RWM-RL & - & $36.70$ & - & - \\
        MAN & - & \underline{$46.53$} & $22.72$ & - \\
        DEBUG & $54.95$ & $37.39$ & $17.69$ & $36.34$ \\
        \hline
        VSLBase & \underline{$61.72$} & $40.97$ & \underline{$24.14$} & \underline{$42.11$} \\
        VSLNet & $\mathbf{64.30}$ & $\mathbf{47.31}$ & $\mathbf{30.19}$ & $\mathbf{45.15}$ \\
        \midrule
        \multicolumn{5}{c}{3D ConvNet with fine-tuning on Charades dataset} \\
        \hline
        ExCL & $65.10$ & $44.10$ & $23.30$ & - \\
        \hline
        VSLBase & \underline{$68.06$} & \underline{$50.23$} & \underline{$30.16$} & \underline{$47.15$} \\
        VSLNet & $\mathbf{70.46}$ & $\mathbf{54.19}$ & $\mathbf{35.22}$ & $\mathbf{50.02}$ \\
        \bottomrule
	\end{tabular}
	\caption{\small Results ($\%$) of ``$\textrm{R@}n, \textrm{IoU}=\mu$'' and ``mIoU'' compared with the state-of-the-art on Charades-STA.}
	\label{tab-sota-charades}
\end{table}

\begin{table}[t]
   \small
	\centering
	\setlength{\tabcolsep}{3.0 pt}
	\begin{tabular}{l c c c c}
		\toprule
        Model & $\textrm{IoU}=0.3$ & $\textrm{IoU}=0.5$ & $\textrm{IoU}=0.7$ & mIoU \\
        \midrule
        TGN & $45.51$ & $28.47$ & -  & - \\
        ABLR & $55.67$ & $36.79$ & - & $36.99$ \\
        RWM-RL & - & $36.90$ & - & - \\
        QSPN & $45.30$ & $27.70$ & $13.60$ & - \\
        ExCL$^{*}$ & \underline{$63.00$} & $\mathbf{43.60}$ & \underline{$24.10$} & - \\
        DEBUG & $55.91$ & $39.72$ & - & $39.51$ \\
        \hline
        VSLBase & $58.18$ & $39.52$ & $23.21$ & $40.56$ \\
        VSLNet & $\mathbf{63.16}$ & \underline{$43.22$} & $\mathbf{26.16}$ & $\mathbf{43.19}$ \\
        \bottomrule
	\end{tabular}
	\caption{\small Results ($\%$) of ``$\textrm{R@}n, \textrm{IoU}=\mu$'' and ``mIoU'' compared with the state-of-the-art on ActivityNet Caption.}
	\label{tab-sota-activitynet}
\end{table}

\begin{table}[t]
   \small
	\centering
	\setlength{\tabcolsep}{3.0 pt}
	\begin{tabular}{l c c c c}
		\toprule
        Model & $\textrm{IoU}=0.3$ & $\textrm{IoU}=0.5$ & $\textrm{IoU}=0.7$ & mIoU \\
        \midrule
        CTRL & $18.32$ & $13.30$ & -  & - \\
        TGN & $21.77$ & $18.90$ & - & - \\
        ACRN & $19.52$ & $14.62$ & - & - \\
        ABLR & $19.50$ & $9.40$ & - & $13.40$ \\
        ACL-K & \underline{$24.17$} & $20.01$ & - & - \\
        L-Net & - & - & - & $13.41$ \\
        SAP & - & $18.24$ & - & - \\
        SM-RL & $20.25$ & $15.95$ & - & - \\
        DEBUG & $23.45$ & $11.72$ & - & $16.03$ \\
        \hline
        VSLBase & $23.59$ & \underline{$20.40$} & \underline{$16.65$} & \underline{$20.10$} \\
        VSLNet & $\mathbf{29.61}$ & $\mathbf{24.27}$ & $\mathbf{20.03}$ & $\mathbf{24.11}$ \\
        \bottomrule
	\end{tabular}
	\caption{\small Results ($\%$) of ``$\textrm{R@}n, \textrm{IoU}=\mu$'' and ``mIoU'' compared with the state-of-the-art on TACoS.}
	\label{tab-sota-tacos}
\end{table}

\begin{table}[t]
   \small
	\centering
    \setlength{\tabcolsep}{2pt}
	\begin{tabular}{l c c c c}
		\toprule
        Module & $\textrm{IoU}=0.3$ & $\textrm{IoU}=0.5$ & $\textrm{IoU}=0.7$ & mIoU \\
        \midrule
        BiLSTM + CAT & $61.18$ & $43.04$ & $26.42$ & $42.83$ \\
        CMF + CAT & $63.49$ & $44.87$ & $27.07$ & $44.01$ \\
        BiLSTM + CQA & $65.08$ & $46.94$ & $28.55$ & $45.18$ \\
        CMF + CQA & $68.06$ & $50.23$ & $30.16$ & $47.15$ \\
        \bottomrule
	\end{tabular}
	\caption{\small Comparison between models with alternative modules in VSLBase on Charades-STA.}
	\label{tab-component}
\end{table}

The results on Charades-STA are summarized in Table~\ref{tab-sota-charades}. For fair comparison with ExCL, we follow the same setting in ExCL to use the 3D ConvNet fine-tuned on Charades dataset as visual feature extractor. Observed that VSLNet significantly outperforms all baselines by a large margin over all metrics. It is worth noting that the performance improvements of VSLNet are more significant under more strict metrics. For instance, VSLNet achieves $7.47\%$ improvement in $\textrm{IoU}=0.7$ versus $0.78\%$ in $\textrm{IoU}=0.5$, compared to MAN. Without query-guided highlighting, VSLBase outperforms all compared baselines over $\textrm{IoU}=0.7$, which shows adopting span-based QA framework is promising for NLVL. Moreover, VSLNet benefits from visual feature fine-tuning, and achieves state-of-the-art results on this dataset.

Table~\ref{tab-sota-activitynet} summarizes the results on ActivityNet Caption dataset. Note that this dataset requires YouTube clips to be downloaded online. We have $1,309$ missing videos, while ExCL reports $3,370$ missing videos. Strictly speaking, the results reported in this table are not directly comparable. Despite that, VSLNet is superior to ExCL with $2.06\%$ and $0.16\%$ absolute improvements over $\textrm{IoU}=0.7$ and $\textrm{IoU}=0.3$, respectively. Meanwhile, VSLNet surpasses other baselines.

\begin{table}[t]
   \small
	\centering
	\begin{tabular}{c c c c}
		\toprule
        Module & CAT & CQA & $\Delta$ \\
        \midrule
        BiLSTM & $26.42$ & $28.55$ & $+2.13$ \\
        CMF & $27.07$ & $30.16$ & $+3.09$ \\
        $\Delta$ & $+0.65$ & $+1.61$ & - \\
        \bottomrule
	\end{tabular}
	\caption{\small Performance gains ($\%$) of different modules over ``$\textrm{R@}1, \textrm{IoU}=0.7$'' on Charades-STA.}
	\label{tab-component-delta}
\end{table}

Similar observations hold on TACoS dataset. Reported in Table~\ref{tab-sota-tacos}, VSLNet achieves new state-of-the-art performance over all evaluation metrics. Without QGH, VSLBase shows comparable performance with baselines.

\begin{figure}[t]
    \centering
	\includegraphics[width=0.48\textwidth]{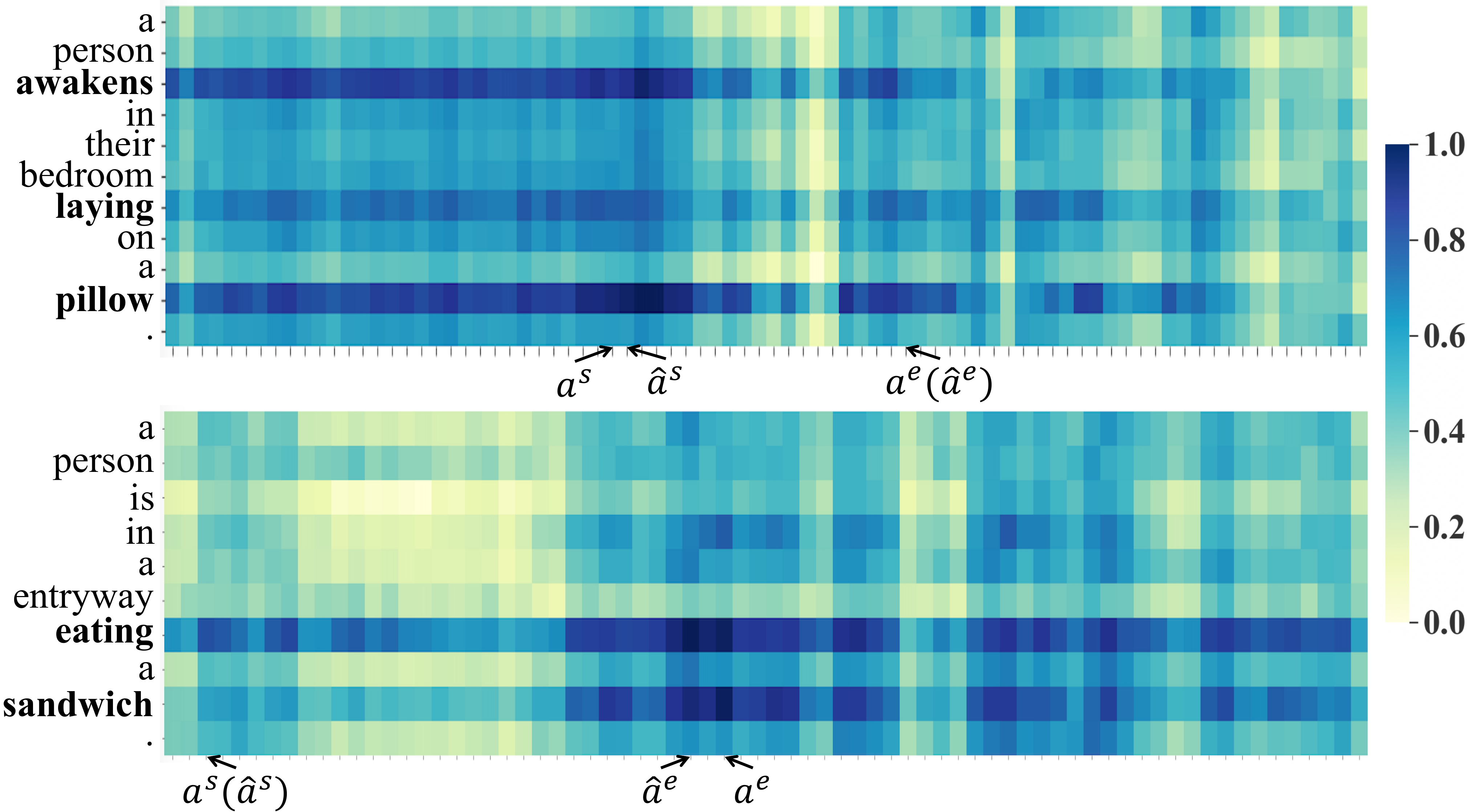}
	\caption{\small Similarity scores, $\mathcal{S}$, between visual and language features in the context-query attention.  $a^s/a^e$ denote the start/end boundaries of ground truth video moment, $\hat{a}^{s}/\hat{a}^{e}$ denote the start/end boundaries of predicted target moment.}
	\label{fig_attn_score}
\end{figure}

\subsection{Ablation Studies}
We conduct ablative experiments to analyze the importance of feature encoder and context-query attention in our approach. We also investigate the impact of extension ratio $\alpha$ (see Figure~\ref{fig_mask}) in query-guided highlighting (QGH). Finally we visually show the effectiveness of QGH in VSLNet, and discuss the weaknesses of VSLBase and VSLNet.

\subsubsection{Module Analysis}
We study the effectiveness of our feature encoder and context-query attention (CQA) by replacing them with other modules. Specifically, we use bidirectional LSTM (BiLSTM) as an alternative feature encoder. For context-query attention, we replace it by a simple method (named CAT) which concatenates each visual feature with max-pooled query feature.

Recall that our feature encoder consists of Convolution + Multi-head attention + Feed-forward layers (see Section~\ref{ssec:feature}), we name it CMF. With the alternatives, we now have 4 combinations, listed in Table~\ref{tab-component}. Observe from the results, CMF shows stable superiority over CAT on all metrics regardless of other modules; CQA surpasses CAT whichever feature encoder is used. This study indicates that CMF and CQA are more effective.

\begin{figure}[t]
    \centering
	\subfigure[\small $\textrm{R@}1, \textrm{IoU}=0.3$]
	{\label{fig_mask_ratio_iou3}	\includegraphics[trim={0.3cm 0.2cm 0.3cm 0.3cm},clip,width=0.23\textwidth]{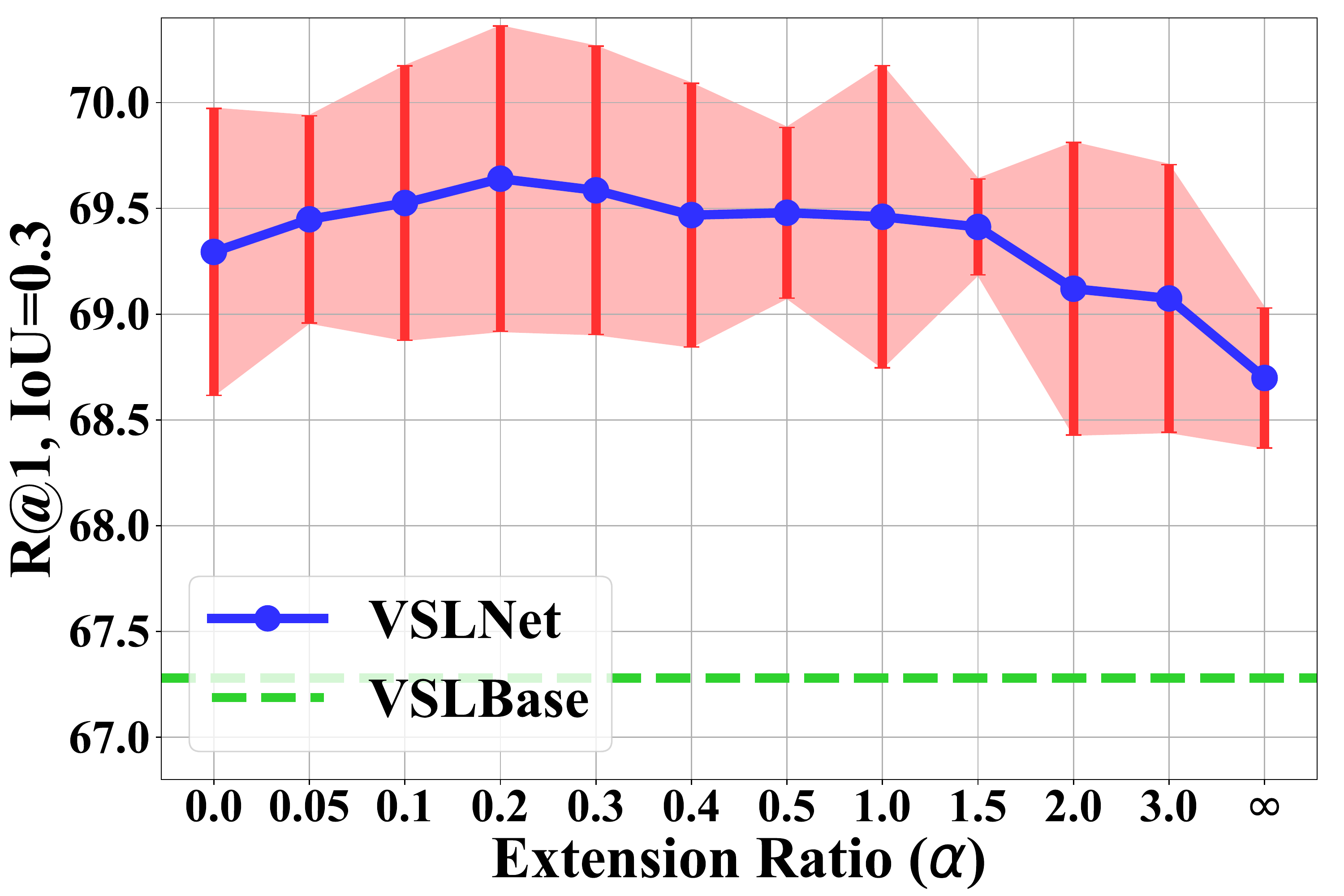}}
	\subfigure[\small $\textrm{R@}1, \textrm{IoU}=0.5$]
	{\label{fig_mask_ratio_iou5}	\includegraphics[trim={0.3cm 0.2cm 0.3cm 0.3cm},clip, width=0.23\textwidth]{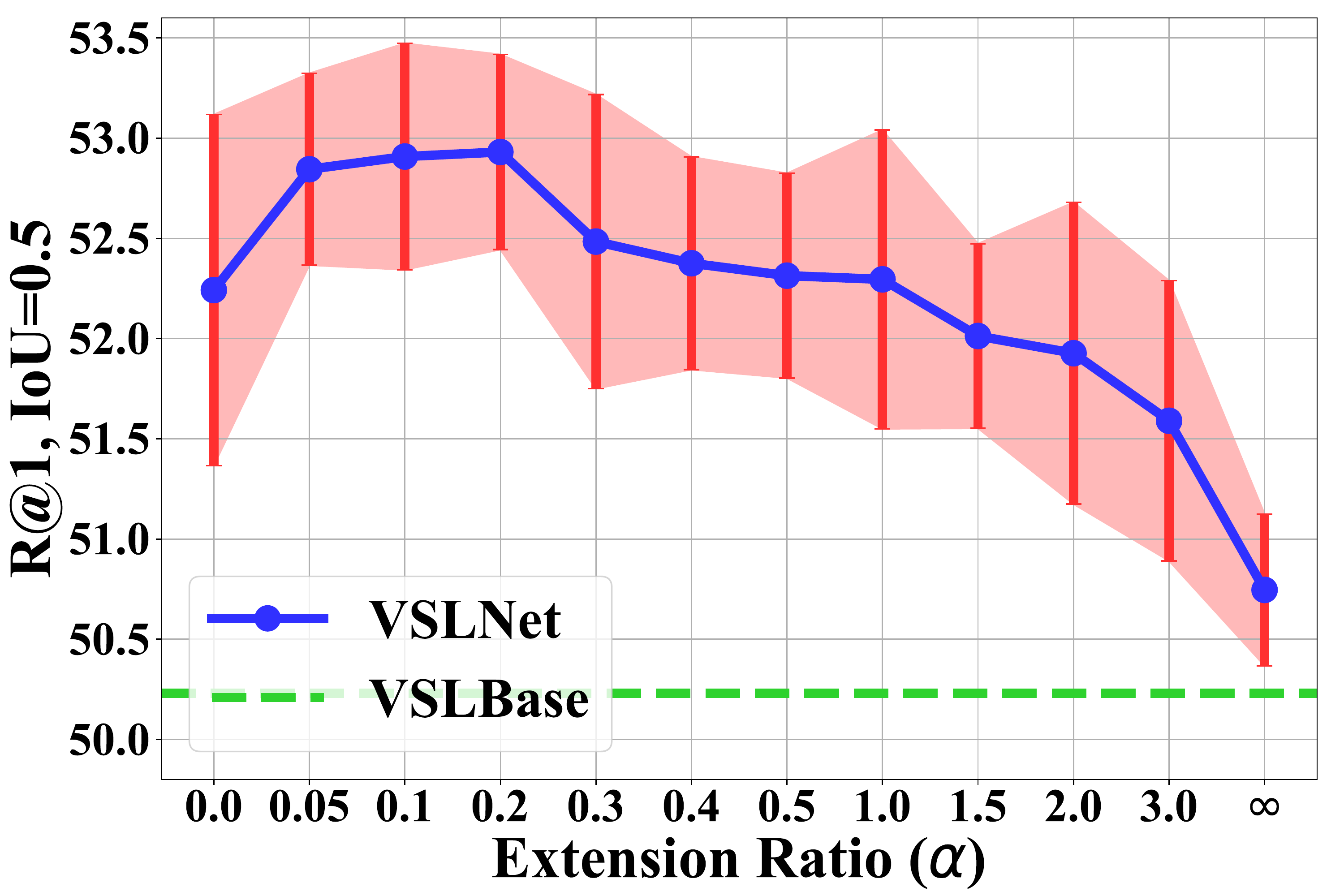}}
	\subfigure[\small $\textrm{R@}1, \textrm{IoU}=0.7$]
	{\label{fig_mask_ratio_iou7}	\includegraphics[trim={0.3cm 0.2cm 0.3cm 0.3cm},clip, width=0.23\textwidth]{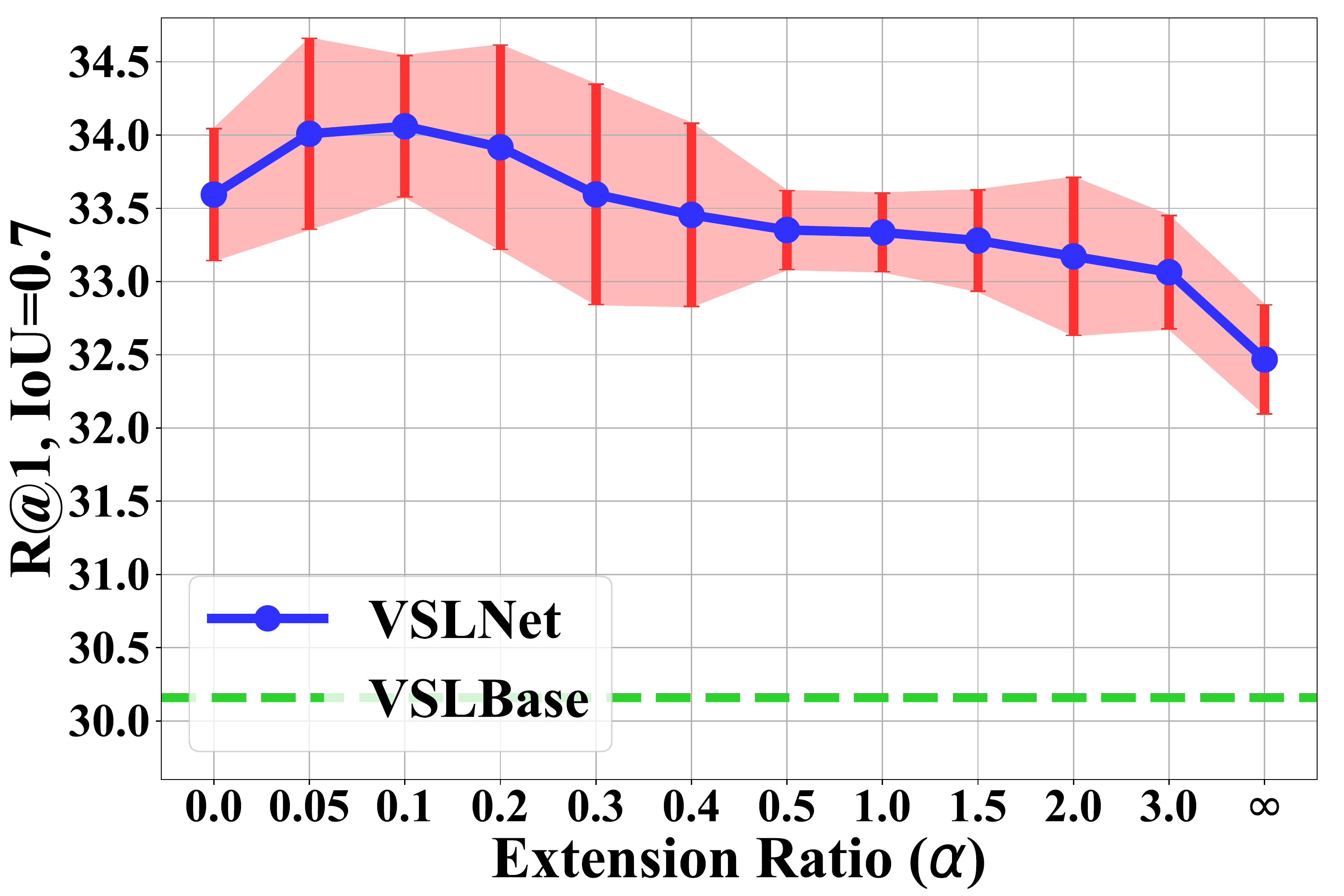}}
	\subfigure[\small mIoU]
	{\label{fig_mask_ratio_miou}	\includegraphics[trim={0.3cm 0.2cm 0.3cm 0.3cm},clip, width=0.23\textwidth]{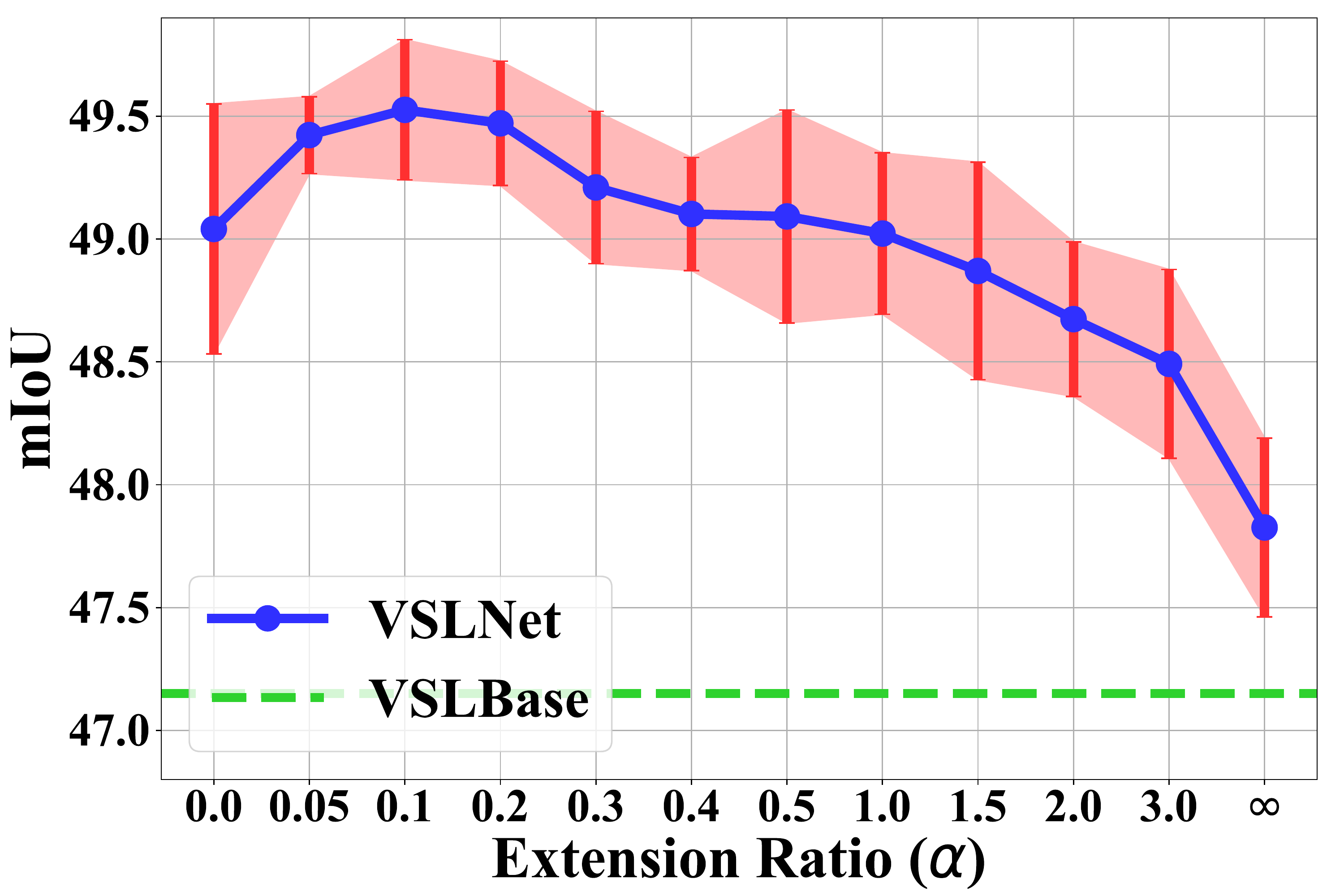}}
	\caption{\small Analysis of the impact of extension ratio $\alpha$ in Query-Guided Highlighting on Charades-STA.}
	\label{fig_charades_mask_ratio}
\end{figure}

\begin{figure}[t]
    \centering
	\subfigure[\small Charades-STA]
	{\label{fig_charades_hist}	\includegraphics[trim={0.3cm 0.2cm 0.3cm 0.3cm},clip, width=0.23\textwidth]{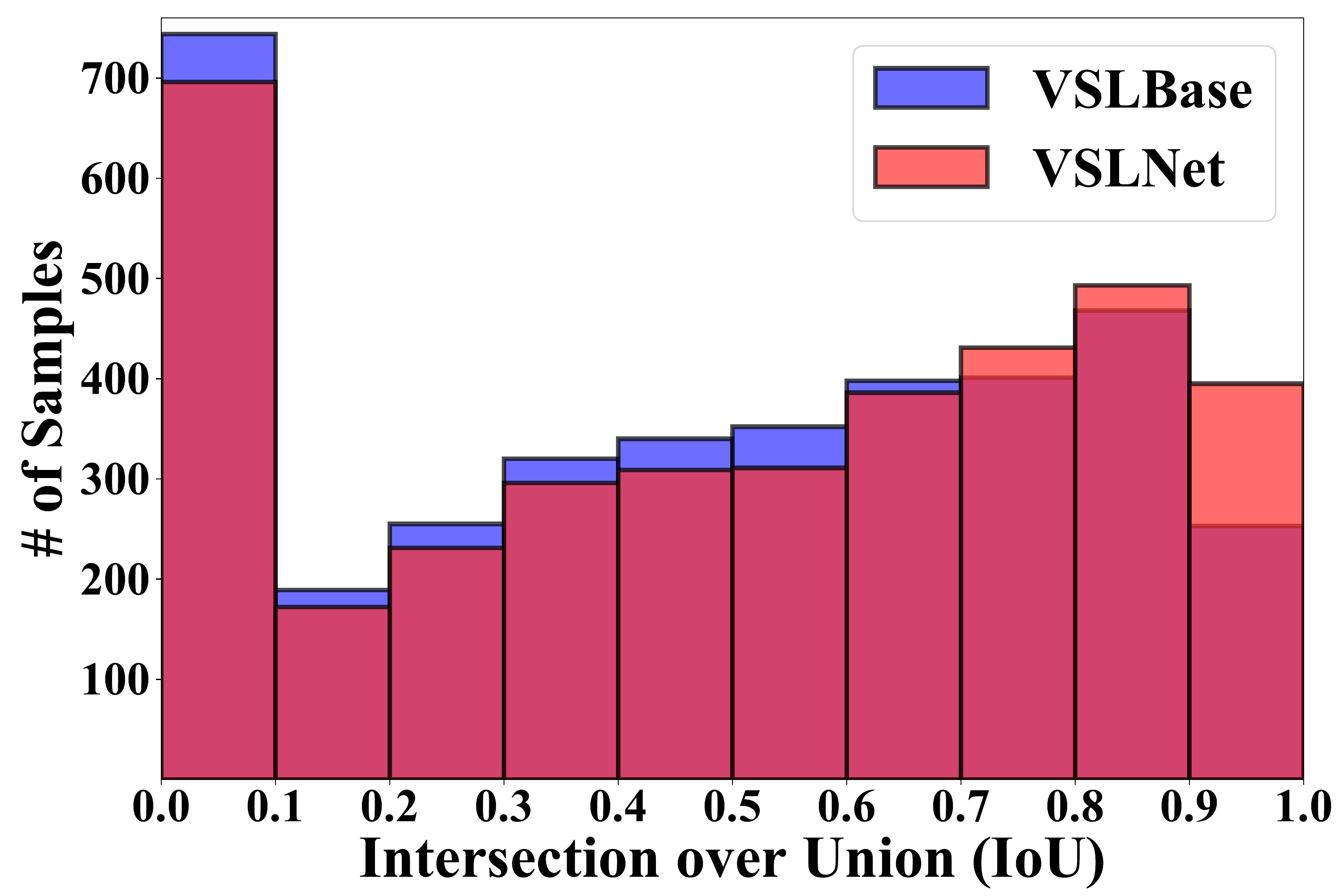}}
	\subfigure[\small ActivityNet Caption]
	{\label{fig_activitynet_hist}	\includegraphics[trim={0.3cm 0.2cm 0.3cm 0.3cm},clip, width=0.23\textwidth]{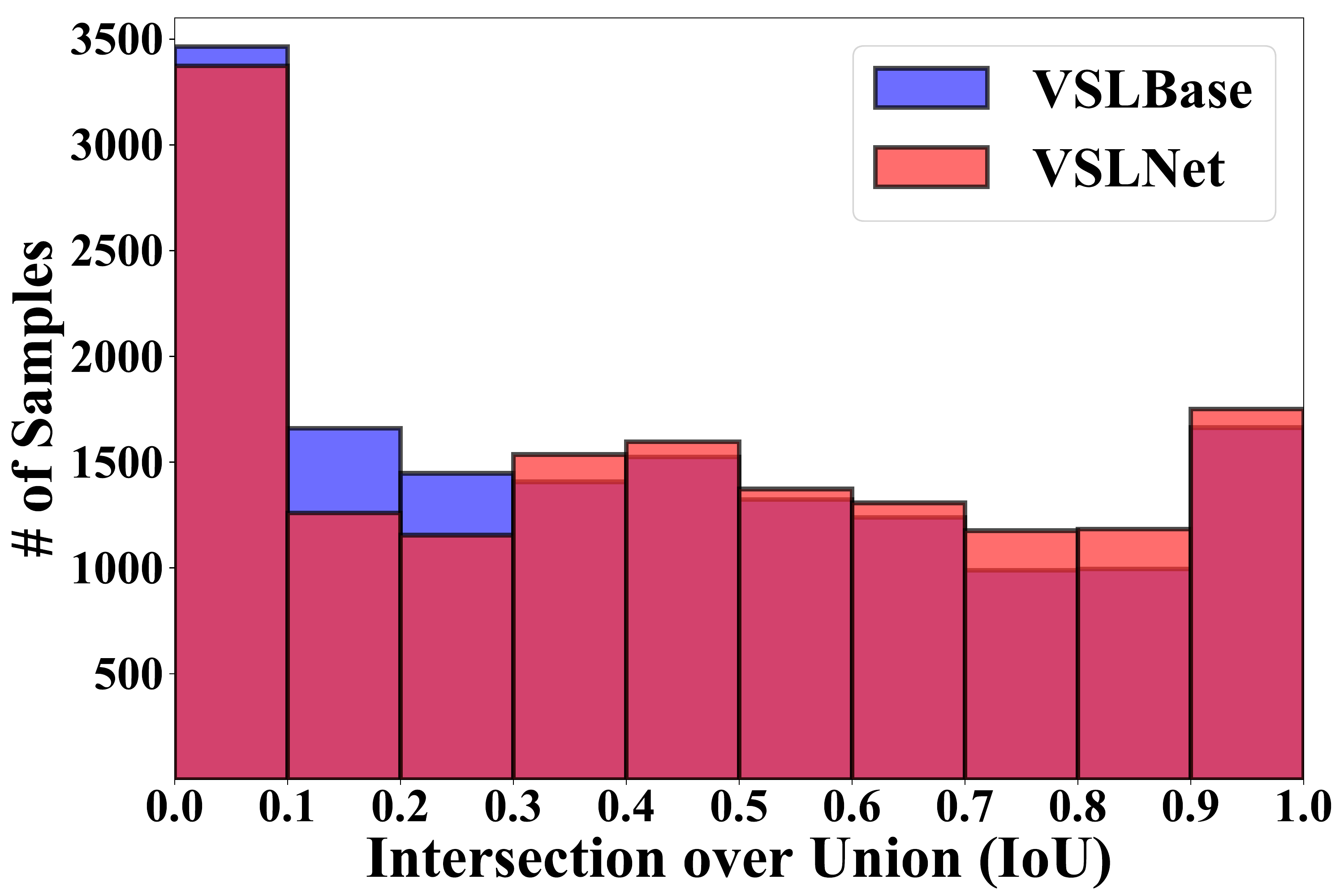}}
	\caption{\small Histograms of the number of predicted results on test set under different IoUs, on two datasets.}
	\label{fig_hist}
\end{figure}

Table~\ref{tab-component-delta} reports performance gains of different modules over ``$\textrm{R@}1, \textrm{IoU}=0.7$'' metric. The results shows that replacing CAT with CQA leads to larger improvements, compared to replacing BiLSTM by CMF. This observation suggests CQA plays a more important role in our model. Specifically, keeping CQA, the absolute gain is $1.61\%$ by replacing encoder module. Keeping CMF, the gain of replacing attention module is $3.09\%$.

Figure~\ref{fig_attn_score} visualizes the matrix of similarity score between visual and language features in the context-query attention (CQA) module ($\mathcal{S}\in\mathbb{R}^{n\times m}$ in Section~\ref{ssec:attention}). This figure shows visual features are more relevant to the verbs and their objects in the query sentence. For example, the similarity scores between visual features and ``\textit{eating}'' (or ``\textit{sandwich}'') are higher than that of other words. We believe that verbs and their objects are more likely to be used to describe video activities. Our observation is consistent with~\citet{ge2019mac}, where  \textit{verb-object} pairs are extracted as semantic activity concepts. In contrast, these concepts are automatically captured by the CQA module in our method.

\begin{figure*}[t]
    \centering
	\subfigure[\small Two example cases on the Charades-STA dataset]
	{\label{fig_qual_charades}	\includegraphics[width=0.95\textwidth]{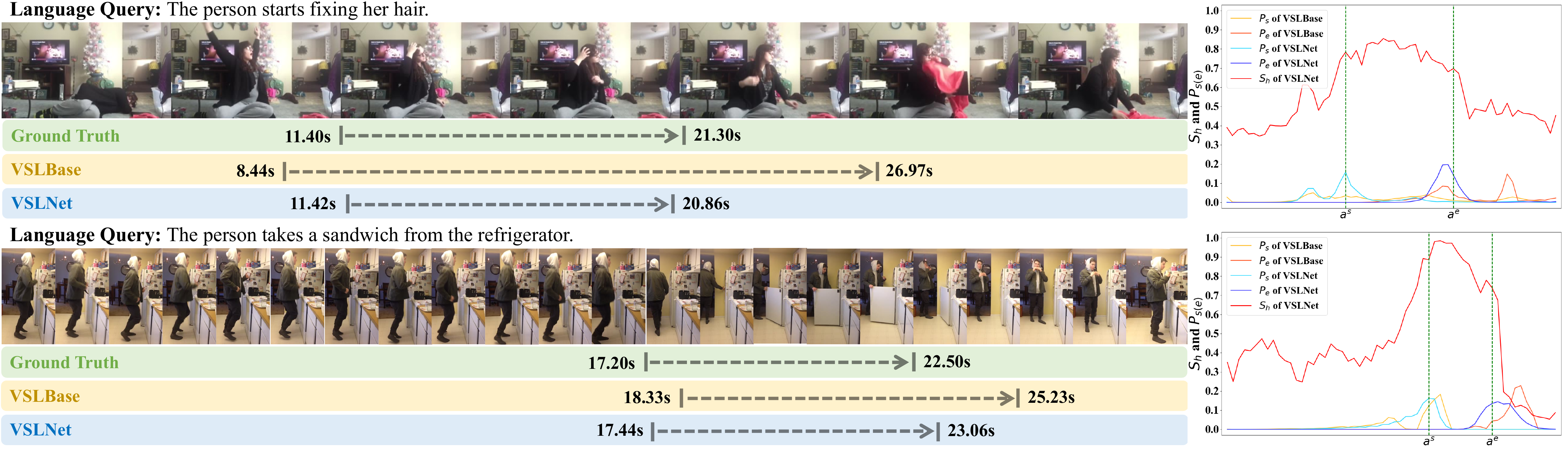}}
	\subfigure[\small Two example cases on the ActivityNet Caption dataset]
	{\label{fig_qual_activitynet}	\includegraphics[width=0.95\textwidth]{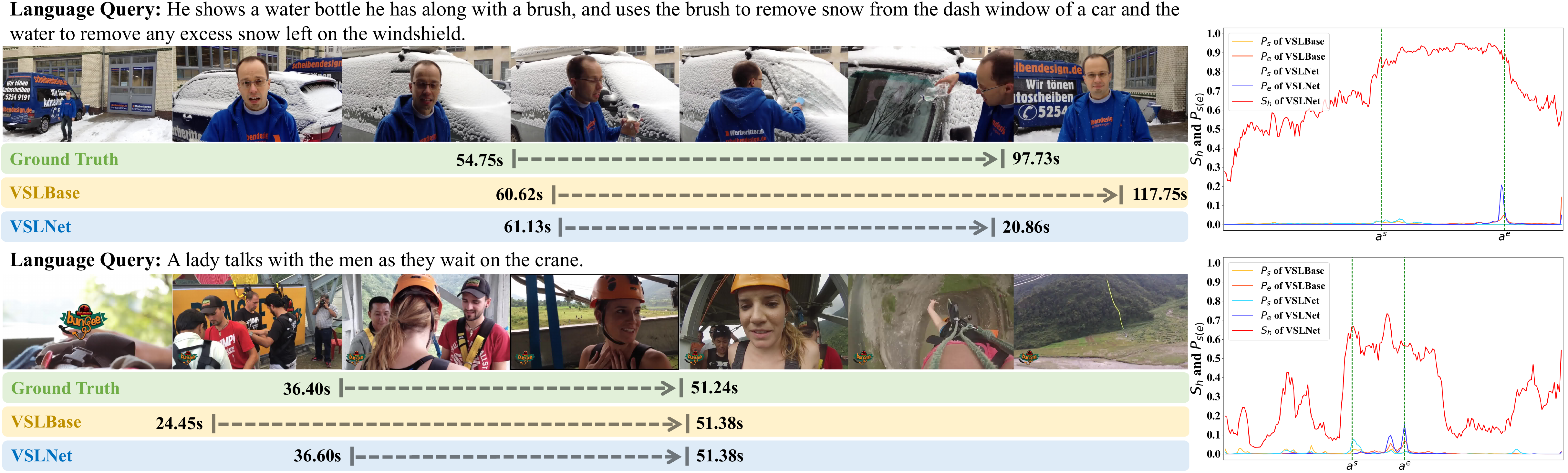}}
	\caption{\small Visualization of predictions by VSLBase and VSLNet. Figures on the left depict the localized results by the two models. Figures on the right show probability distributions of start/end boundaries and highlighting scores.}
	\label{fig_qual}
\end{figure*}

\begin{figure}[t]
    \centering
	\subfigure[\small Charades-STA]
	{\label{fig_charades_error_time_hist}	\includegraphics[trim={0cm 0cm 0cm 0cm},clip, width=0.23\textwidth]{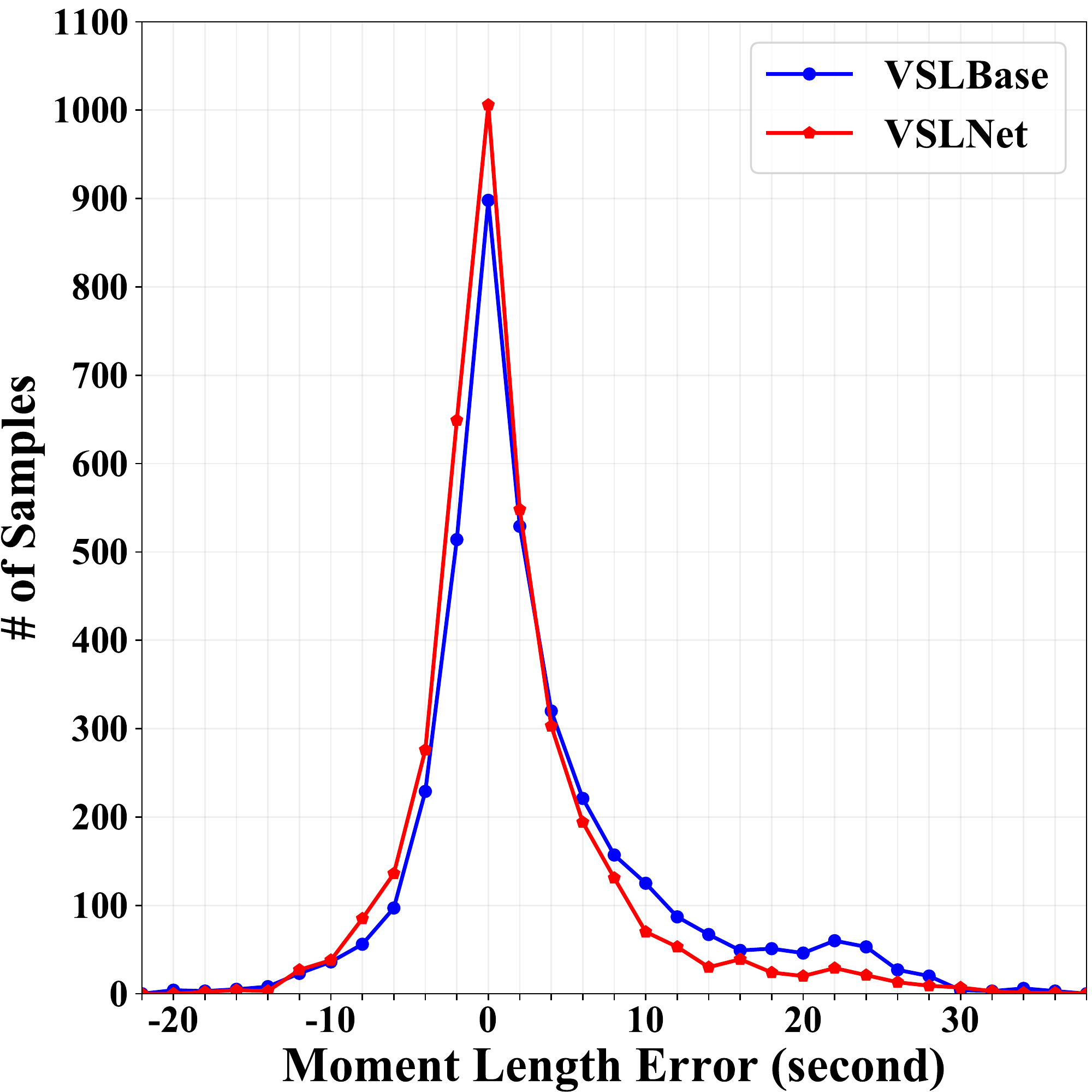}}
	\subfigure[\small ActivityNet Caption]
	{\label{fig_activitynet_error_time_hist}	\includegraphics[trim={0cm 0cm 0cm 0cm},clip, width=0.23\textwidth]{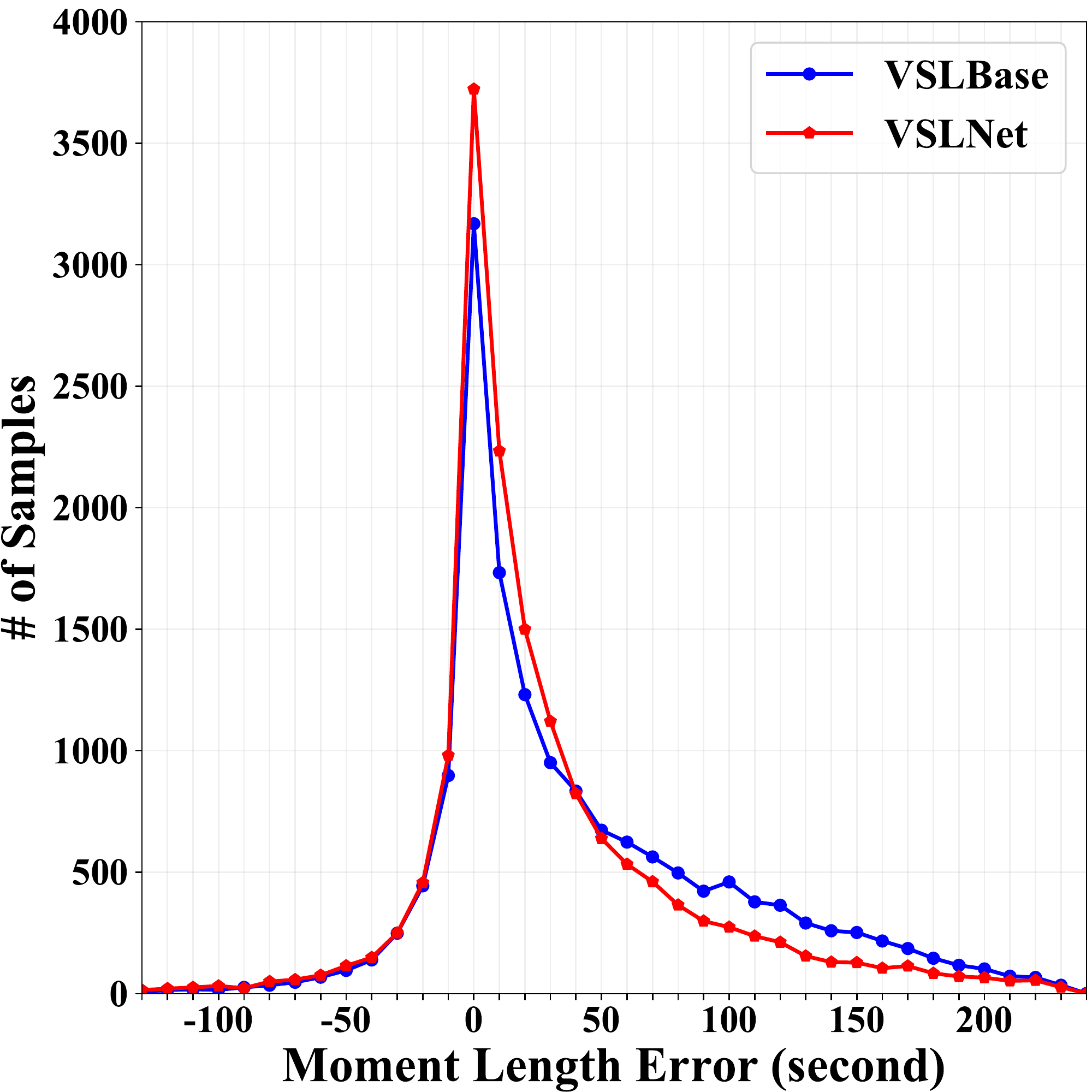}}
	\caption{\small Plots of moment length errors in seconds between ground truths and results predicted by VSLBase and VSLNet, respectively.}
	\label{fig_length_diff_hist}
\end{figure}

\subsubsection{The Impact of Extension Ratio in QGH}
We now study the impact of extension ratio $\alpha$ in query-guided highlighting module on Charades-STA dataset. We evaluated $12$ different values of $\alpha$ from $0.0$ to $\infty$ in experiments. $0.0$ represents no answer span extension, and $\infty$ means that the entire video is regarded as foreground.

The results for various $\alpha$'s are plotted in Figure~\ref{fig_charades_mask_ratio}. It shows that query-guided highlighting consistently contributes to performance improvements, regardless of $\alpha$ values, \ie from $0$ to $\infty$. 

Along with $\alpha$ raises, the performance of VSLNet first increases and then gradually decreases. The optimal performance appears between $\alpha=0.05$ and $0.2$ over all metrics. 

Note that, when $\alpha=\infty$, which is equivalent to no region is highlighted as a coarse region to locate target moment, VSLNet remains better than VSLBase. Shown in Figure~\ref{fig_qgm}, when $\alpha=\infty$, QGH effectively becomes a straightforward concatenation of sentence representation with each of visual features. The resultant feature remains helpful for capturing semantic correlations between vision and language. In this sense, this function can be regarded as an approximation or simulation of the traditional multimodal matching strategy~\cite{Hendricks2017LocalizingMI,Gao2017TALLTA,Liu2018AMR}.

\subsubsection{Qualitative Analysis}
Figure~\ref{fig_hist} shows the histograms of predicted results on test sets of Charades-STA and ActivityNet Caption datasets. Results show that VSLNet beats VSLBase by having more samples in the high IoU ranges, \eg $\textrm{IoU}\geq 0.7$ on Charades-STA dataset. More predicted results of VSLNet are distributed in the high IoU ranges for ActivityNet Caption dataset. This result demonstrates the effectiveness of the query-guided highlighting (QGH) strategy.

We show two examples in Figures~\ref{fig_qual_charades} and~\ref{fig_qual_activitynet} from Charades-STA and ActivityNet Caption datasets, respectively. From the two figures, the localized moments by VSLNet are closer to ground truth than that by VSLBase. Meanwhile, the start and end boundaries predicted by VSLNet are roughly constrained in the highlighted regions $S_{\textrm{h}}$, computed by QGH.

\begin{figure*}[t]
    \centering
	\subfigure[\small A failure case on the Charades-STA dataset with $\textrm{IoU}=0.11$.]
	{\label{fig_qual_charades_error}	\includegraphics[width=0.95\textwidth]{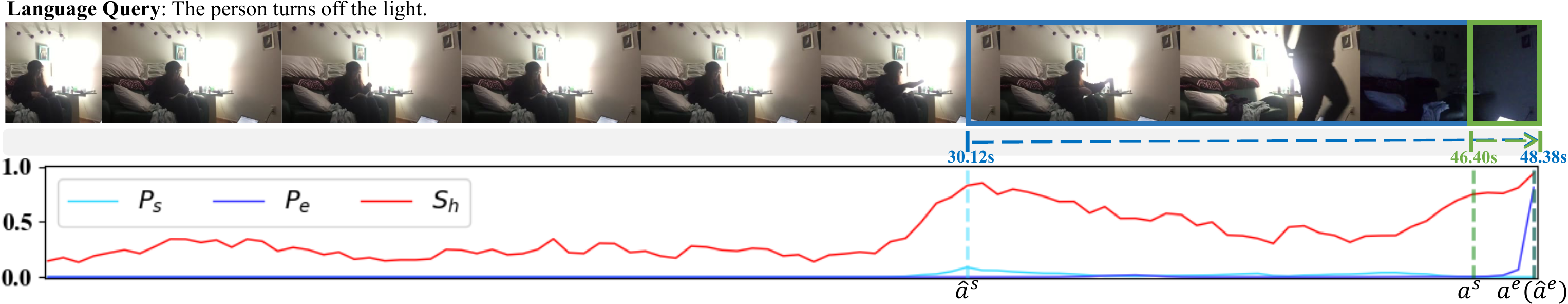}}
	\subfigure[\small A failure case on the ActivityNet Caption dataset with $\textrm{IoU}=0.17$.]
	{\label{fig_qual_activitynet_error}	\includegraphics[width=0.95\textwidth]{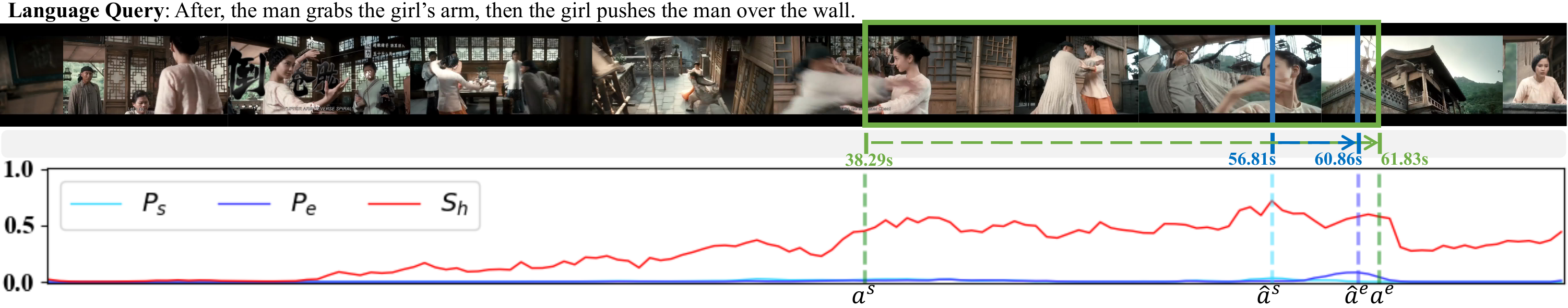}}
	\caption{\small Two failure examples predicted by VSLNet, $a^s/a^e$ denote the start/end boundaries of ground truth video moment, $\hat{a}^{s}/\hat{a}^{e}$ denote the start/end boundaries of predicted target moment.}
	\label{fig_qual_error}
\end{figure*}

We further study the error patterns of predicted moment lengths, as shown in Figure~\ref{fig_length_diff_hist}. The differences between moment lengths of ground truths and predicted results are measured. A positive length difference means the predicted moment is longer than the corresponding ground truth, while a negative means shorter. Figure~\ref{fig_length_diff_hist} shows that VSLBase tends to predict longer moments, \eg more samples with length error larger than $4$ seconds in Charades-STA or 30 seconds in ActivityNet. On the contrary, constrained by QGH, VSLNet tends to predict shorter moments, \eg more samples with length error smaller than $-4$ seconds in Charades-STA or $-20$ seconds in ActivityNet Caption. This observation is helpful for future research on adopting span-based QA framework for NLVL.

In addition, we also exam failure cases (with IoU predicted by VSLNet lower than $0.2$) shown in Figure~\ref{fig_qual_error}. In the first case, as illustrated by Figure~\ref{fig_qual_charades_error}, we observe an action that a person turns towards to the lamp and places an item there. The QGH falsely predicts the action as the beginning of the moment "turns off the light". The second failure case involves multiple actions in a query, as shown in Figure~\ref{fig_qual_activitynet_error}. QGH successfully highlights the correct region by capturing the temporal information of two different action descriptions in the given query. However, it assigns ``pushes'' with higher confidence score than ``grabs''. Thus, VSLNet only captures the region corresponding to the ``pushes'' action, due to its confidence score.

\section{Conclusion}
By considering a video as a text passage, we solve the NLVL task with a multimodal span-based QA framework. Through experiments, we show that adopting a standard span-based QA framework, VSLBase, effectively addresses NLVL problem. However, there are two major differences between video and text. We further propose VSLNet, which introduces a simple and effective strategy named query-guided highlighting, on top of VSLBase. With QGH, VSLNet is guided to search for answers within a predicted coarse region. The effectiveness of VSLNet (and even VSLBase) suggest that it is promising to explore span-based QA framework to address NLVL problems.

\section*{Acknowledgments}
This research is supported by the Agency for Science, Technology and Research (A*STAR) under its AME Programmatic Funding Scheme (Project \#A18A1b0045 and \#A18A2b0046).

\bibliography{acl2020}
\bibliographystyle{acl_natbib}

\end{document}